\documentclass[acmtog]{acmart}

\setcopyright{none}

\setcitestyle{square}

\usepackage{multirow}

\begin{document}

\title{Unit Region Encoding: A Unified and Compact Geometry-aware Representation for Floorplan Applications}

\author{Huichao Zhang}
\email{zhanghuichao.hc@bytedance.com}
\affiliation{%
 \institution{ByteDance}
   \country{China}}

\author{Pengyu Wang}
\affiliation{%
 \institution{Alibaba}
    \country{China}}

\author{Manyi Li}
\affiliation{%
 \institution{Shandong University}
   \country{China}}

\author{Zuojun Li}
\affiliation{%
 \institution{Alibaba}
   \country{China}}

\author{Yaguang Wu}
\affiliation{%
 \institution{Alibaba}
   \country{China}}



\begin{abstract}
We present the Unit Region Encoding of floorplans, which is a unified and compact geometry-aware encoding representation for various applications, ranging from interior space planning, floorplan metric learning to floorplan generation tasks. The floorplans are represented as the latent encodings on a set of boundary-adaptive unit region partition based on the clustering of the proposed geometry-aware density map. The latent encodings are extracted by a trained network (URE-Net) from the input dense density map and other available semantic maps. Compared to the over-segmented rasterized images and the room-level graph structures, our representation can be flexibly adapted to different applications with the sliced unit regions while achieving higher accuracy performance and better visual quality. We conduct a variety of experiments and compare to the state-of-the-art methods on the aforementioned applications to validate the superiority of our representation, as well as extensive ablation studies to demonstrate the effect of our slicing choices. 
\end{abstract}

\ccsdesc[500]{Computing methodologies~Computer graphics}
\ccsdesc[300]{Computing methodologies~Neural networks}
\ccsdesc[300]{Computing methodologies~Shape representations}
\ccsdesc[300]{Theory of computation~Computational geometry}

\keywords{floorplan representation, geometric shape, neural networks, semantic segmentation, structural Similarity, floorplan generation}

\begin{teaserfigure}
  \includegraphics[width=\textwidth]{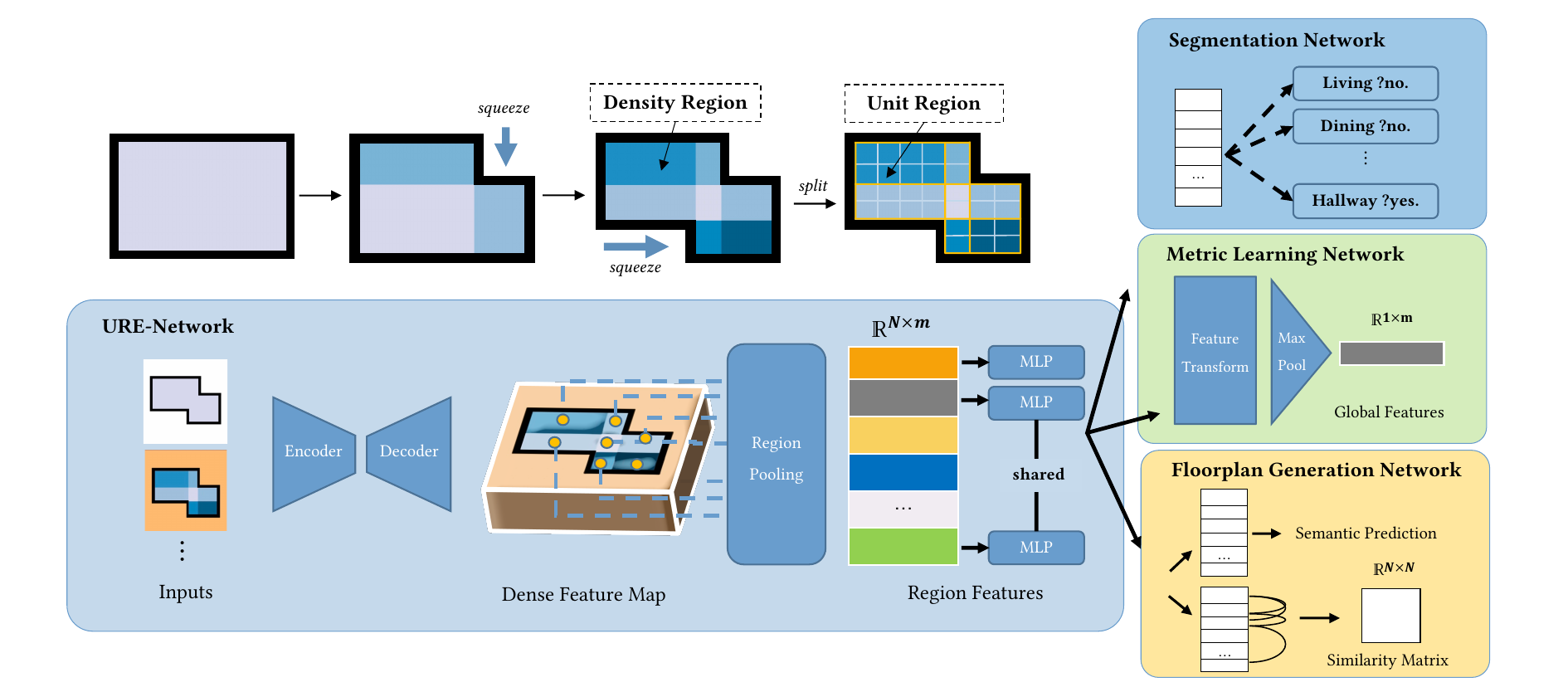}
  \caption{We present the Unit Region Encoding of floorplans, which is a unified and compact geometry-aware encoding representation for various applications. As shown in the top row, we consider the floorplan as produced by a sequence of squeeze and splitting operations, which naturally forms the unit region partition. We propose URE-Net (bottom row) to learn the region-wise encoding, which can be flexibly adapted to many different applications.}
  \Description{}
  \label{fig:teaser}
\end{teaserfigure}

\maketitle

\section{Introduction}

\begin{figure*}[h]
  \centering
  \includegraphics[width=\linewidth]{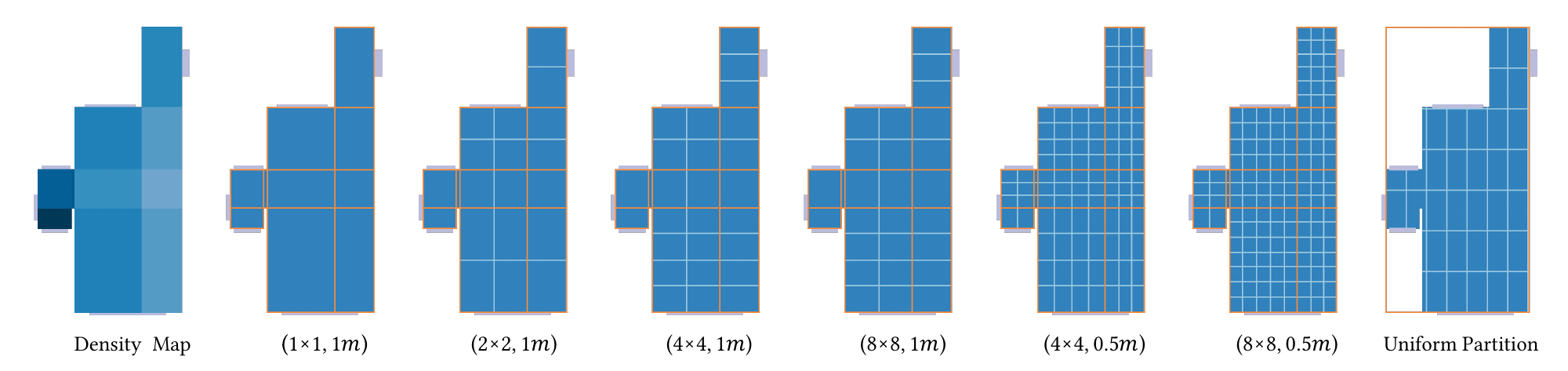}
  \caption{The visualization of the proposed density map and unit region partition under different splitting strategies. The last picture shows the uniform partition without considering the particular floorplan shape, causing the misalignment between the regions and the floorplan boundary and thus the irregular regions.}
  \Description{ttttttttttt.}
  \label{fig:unit_region}
\end{figure*}

Floorplans play an essential role in architecture designs and analysis to indicate the room structures in the form of 2D layouts. The professional softwares such as AutoCAD enable the user to draw and visualize the floorplans in a vector-graphics representation\cite{kumar2017assisting}. In the related research area of computer graphics, the floorplans are converted to various data structures, e.g. binary trees~\cite{Yao2003FloorplanRC}, SVG format\cite{carlier2020deepsvg}, parametric representation~\cite{Wu2018MIQPbasedLD}, for different tasks including floorplan design, organization, indoor scene synthesis, and path planning, etc. \cite{hu2020graph2plan, patil2021layoutgmn,zhang2020deep,xu2017bim}.


The need for an appropriate and unified representation becomes more urgent with the increasing interest in the data-driven floorplan applications. The most popular choice is to convert the floorplans into 2D rasterized images by quantifying the rich semantic annotations as image channels \cite{wu2019data,wang2018deep,ritchie2019fast,wang2021actfloor}. The input is often represented as a multi-channel image including the masks of interior space, exterior walls and doors, etc, while the output is a global feature describing the entire floorplan or a pixel-wise segmentation labels.
Although this representation fits well with the dominant 2D convolution neural networks,
the overlook of the structural space partition of the floorplans causes the limited accuracy performance in the related applications as pointed in~\cite{patil2021layoutgmn,azizi2021graph} and the low-quality segmentation results with bumpy boundary and some isolated regions. Recently there is a growing trend to convert the floorplans into the structural graph representations\cite{sharma2016unified, sharma2018high, azizi2021graph, patil2021layoutgmn}, where the nodes for the rooms and edges for the adjacency relations. The graph neural networks are then used for the floorplan encoding. However, the nodes of the input graphs are just some simple hand-crafted geometric features such as room sizes and areas with the detailed geometric shape information missing, which affects the accuracy performance of some applications. 


In this work, we propose the Unit Region Encoding (URE) as a unified and compact geometry-aware encoding representation of floorplans. The key motivation is that the complex floorplans should be encoded on a carefully-designed region partition to be flexibly adapted to various applications. Based on the observation that the semantic information of floorplans are highly aligned with the irregular boundaries, i.e. the interior walls 
dividing the rooms or functional areas are often aligned with some segments of the exterior boundaries, we represent the floorplan as a grid of boundary-adaptive unit regions as in Figure~\ref{fig:teaser}, each with a latent encoding feature. Specifically, we define the density map where the density of each location is defined as the sum of nearest distances along the four directions to the floorplan boundaries, then perform the clustering and splitting to obtain the unit region partition. We train our URE-Net on the floorplan image and the density map to obtain the pixel-wise dense feature map and integrate to obtain the region-wise encoding features.

Let's examine the correlation between our unit region partition and the floorplan structures by assigning the semantic room labels to the regions, i.e. all the pixels in each region should maintain the room label as the majority of them. We tested the IoU accuracy under different splitting strategies by controlling the parameters of the unit region partition. As shown in Table~\ref{tab:correlation_check}, Our $(8\times8\times1m)$ and $(8\times8\times0.5m)$ settings achieve the IoU accuracy up to $99.10\%$ and $99.98\%$ on the 3D-Front dataset~\cite{fu20213d}, as well as $93.47\%$ and $95.99\%$ on the RPLAN dataset~\cite{wu2019data}. Comparing to the $256\times256$ rasterized images, our partition is able to represent the complex room structures with hundreds times less of the units with negligible loss on the accuracy.

\begin{table}[tp]
   \fontsize{5.5}{8}\selectfont  
   \centering  
  \caption{The IoU accuracy and average region numbers of the unit region partition to represent the floorplans, under different splitting strategies. The last two rows are the measurements of the dense $256\times256$ image representation, and the uniform partition regardless of the floorplan shape.}
  \label{tab:correlation_check}
  \begin{tabular}{lcc|cc|cc}

    \toprule
    \multirow{3}*{Strategy} & \multirow{3}*{Grid} & \multirow{3}*{Thresh} &
    \multicolumn{2}{c|}{3D-FRONT~\cite{fu20213d}} & \multicolumn{2}{c}{RPLAN~\cite{wu2019data}} \\
    \cline{4-7}
     {} & {} & {} &IoU$(\uparrow)$ & Avg. Num$(\downarrow)$ & IoU$(\uparrow)$ & Avg. Num$(\downarrow)$ \\
    \midrule
    \textbf{[1]Density Region} & 1$\times$1 & 1.0m & 95.28&8.76&65.42&14.26\\
    \textbf{[2]Unit Region} & 2$\times$ 2& 1.0m& 97.83&19.33&81.40&36.52\\
    \textbf{[3]Unit Region} & 4$\times$ 4& 1.0m& 99.40&34.66&90.52&71.15\\
    \textbf{[4]Unit Region} & 8$\times$ 8& 1.0m& 99.10&40.41&93.47&88.72\\
    \textbf{[5]Unit Region} & 4$\times$ 4& 0.5m& 99.39&66.35&91.32&133.24\\
    \textbf{[6]Unit Region} & 8$\times$ 8& 0.5m& 99.98&124.89&95.99&266.53\\
    \textbf{[7]Pixel-256} & - & - & 100&24137.18    &100 &17075.31    \\
    \textbf{[8]Uniform Partition} &-&-& 77.20 & 40.74 & 87.81 & 87.48\\
    \bottomrule
  \end{tabular}
\end{table}


Overall, our representation is \emph{compact} as defined on the unit region partition and is \emph{geometry-aware} as the density map is involved in both the partition process and encoding process, which makes it a \emph{unified} representation for many applications.
For the synthesis tasks to generate the structural partitions, such as the interior space planning, our representation significantly reduces the learning space and results in more accurate and neat room partitions. On the other hand for the classification or retrieval task, the unit regions can be easily grouped to the given room partition and the encoding feature per region are integrated to represent the rich geometric information. 
We conduct extensive experiments to validate the superiority of our representation in the applications ranging from interior space planning, floorplan metric learning, and a more challenging floorplan generation task, by comparing with the state-of-the-art methods. 
We also study the effect of density map and the different splitting strategies of the unit region partition to understand the importance of the key designs in our approach.

\section{Related Work}

\textbf{Floorplan Representation.} Floorplans are generally organized in vector-graphics representation to represent boundaries of rooms or houses. Earlier works focus on physical modeling methods \cite{arvin2002modeling, xu2002constraint}, or grammar based optimization \cite{merrell2011interactive,rodrigues2013evolutionary,qi2018human} for scene synthesis or floorplan generation, where the floorplan shape is directly modeled as physical element or constraint. 

The recent trend is to use the data-driven methods to learn expressive features from the large-scale datasets. These methods generally rasterize the floorplans into binary mask images for the convinience to learn with 2D convolution neural networks~\cite{wu2019data,wang2018deep,ritchie2019fast,wang2021actfloor}. However, the binary mask is short of explicit geometric features, and the convolution operation is not guaranteed to preserve the sharp corners and edges
\cite{girard2021polygonal,li2019topological,zorzi2021machine}. On the other hand, some works convert the floorplan into structural graph for high-level representation learning tasks, such as floorplan retrieval \cite{patil2021layoutgmn, azizi2021graph} and floorplan generation \cite{nauata2020house, nauata2021house, para2021generative}. Though being more aware of the floorplan structures, the nodes and edges in the graph are often defined as some simple hand-craft local geometric features, such as room sizes and areas, with the detailed geometric shape information missing. In contrast, our proposed representation learns the latent encoding features for a set of unit regions from the floorplan image with the proposed geometry-aware normalized density map, which combines the advantages of the above two aspects and leads to better performance on a variety of applications.

\textbf{Interior space planning.} 
Interior space planning is the fundamental operation of interior design process which allocates the room space to meet the goal of human activities\cite{2018A, mitton2021residential}. Human designers draw up plans to define different functional zones according to the floorplan and finally add details of furnitures placement. This task would be benificial for scene synthesis\cite{li2019grains,paschalidou2021atiss}, especially for complicated floorplans.

The space planning can be considered as the semantic segmentation task that holds each semantic space in neat and regular regions. Previous semantic segmentation models and variants \cite{chen2018encoder,ronneberger2015u,sun2019deep} show great performance on several segmentation benchmarks\cite{Everingham10,Cordts2016Cityscapes} based on fully convolutional neural networks and multi-scale features. When performing on floorplans, these methods usually lead to blurred boundaries and fragmented results, which is hard to be vectorized into a neat interior space planning without losing the accuracy. 
RNNs such as PolyMapper\cite{li2019topological, castrejon2017annotating} sequentially predict the boundary points by exploiting a CNN-RNN network. But the recurrent structure often produces limited numbers of vertices, making it hard to represent the complex segmentation of floorplans. DefGrid\cite{gao2020beyond} propose a deformable grid structure that can adaptively predict position offsets of the boundary for semantic tasks. The significantly reduced grid resolution leads to more efficient and accurate results. However, the offsets of the grid vertices are hard to learn due to the absence of distinguishable features in input binary floorplan masks. 

\textbf{Floorplan metric learning.}
The comparison between floorplans, especially the complicated floorplans with multiple rooms and irregular boundaries are challenging. Earlier approaches utilize pre-defined image features for comparing the similarity between flooplans, such as Run-Length Histogram\cite{de2013runlength}. In \cite{sharma2017daniel}, a deep network was proposed to learn the embedding vectors of floorplans for the goal of floorplan matching. Grpah structures \cite{wessel2008room, sharma2016unified} express the structural room layouts of floorplans more accurately and show better performance. Recently, LayoutGMN\cite{patil2021layoutgmn} take a predominantly structural view of layouts for both data representation and layout comparison. The proposed attention-based graph matching network performs intra-graph message passing and cross-graph information communication to learn the room structures. However these approaches require hand-craft features for graph nodes and edges, the complex geometrical shape information of floorplans are not fully employed. 

\textbf{Floorplan generation.}
Designing a floorplan means to determine the room partition and functionality labels depending on given exterior boundary of a building and other constraints. 
It is often formulated and solved as a non-deterministic floorplan generation problem\cite{merrell2010computer,wu2019data,hu2020graph2plan,wang2021actfloor,para2021generative}. The early efforts focused more on the optimization-based methods, such as evolutionary algorithms\cite{rodrigues2013approach}, integer programming\cite{Wu2018MIQPbasedLD} and constrained optimization based on graphs\cite{para2021generative}. Recently \cite{wu2019data} proposed a data-driven method to sequentially predict room locations and walls by four convolution neural networks trained on the proposed large-scale residential floorplan dataset RPLAN. Finally they used a post-processing step to convert the wall map into a vector representation. The Grpah2Plan method\cite{hu2020graph2plan} retrieves a layout graph that encodes room attributes and inter-room connections from dataset to assist the generation process. The graph and building boundary are feed into GNN and CNN models to output initial room bounding boxes, then processed by a refine network and vectorization process. Apart from the graph-based approaches, the GAN-based methods are proposed to generate diverse and plausible floorplan images\cite{nauata2020house,nauata2021house,wang2021actfloor}. However, all the above data-driven approaches require lengthy steps for generation and tedious post-processing for vectorization. 
\section{Floorplan Representation}

The main challenge of the data-driven floorplan representations is that the floorplans are often of highly irregular shapes with complex boundaries. In our approach, we define the geometry-aware density map on the floorplan, then perform clustering and splitting to divide the floorplan into a set of unit regions. We propose a URE-Net to learn the latent encodings on the compact unit region partition, which can be widely used in different applications.


\subsection{Density Map}

We assume that each floorplan with irregular shape is obtained from a base floorplan with an axis-aligned bounding box shape by a sequence of "squeeze" operations, see the top row of Figure~\ref{fig:teaser}. The squeeze operation makes an inward push of the boundary and thus causes the density change of the inner space. Therefore, we develop the concept of density map to describe the geometric feature of the boundary shape.





The density map describes the degree of tightness at each location after the squeeze operations. We derive the definition of density map from the concept of Signed Distance Function (SDF) which has been widely used to represent geometry data\cite{curless1996volumetric,chan2005level,park2019deepsdf}. Specifically, the density value at each location is defined as the inverse of the sum of nearest distances to the boundary along the four directions, i.e. north, south, east and west directions:
\begin{equation}
  f(x)=\left\{
    \begin{aligned}
      &\frac{1}{\sum_{\bold v \in \bold V}d(x, \partial\Omega, \bold v)} & \ \rm{if}  x \in \Omega \ &\\
      &\ 0 & \rm{otherwise} & ,
    \end{aligned}
  \right.
\end{equation}
\begin{equation}
  d(x, \partial\Omega, \bold v):=\mathop{inf}\limits_{y\in \partial\Omega^{\bold v}_x}d(x, y) \ ,
\end{equation}
where $x$ is an arbitrary point location on the 2D layout, $\bold V$ is the set of four directions, $\Omega$ is the interior space of the floorplan,  and $\partial\Omega ^{\bold v}_x$ denotes the set of all the points on the floorplan boundary and are on the corresponding direction $\bold v$ to the point $x$. In our implementation, we search the nearest points on the boundary along the four directions and sum up the distances. 

Taking the floorplan in Figure~\ref{fig:teaser} as example, the base floorplan has uniform density distribution within the inner space. After the first squeeze operation 
at the corner of the floorplan boundary, the change of the density naturally forms three regions inside the floorplan. The points in a more narrow region has higher density values. Similarly, the following squeeze operations further modify the density map and divide the floorplan into finer regions, which enables us to construct the unit region partition.

\subsection{Unit Region Partition}

We divide the floorplan into a set of unit regions based on the density map. Specifically, we first cut the inner space of the floorplan into several \emph{density regions} via a clustering process, which groups the connected locations with the same density value into the same region. In addition, for the sloping walls or the curved walls which are not horizontal or vertical, we take the regions connected by a piece of continuous wall into one density region.

The \emph{density regions} need to be further sliced into finer grids with specified splitting strategy to construct the \emph{unit region} partition, which increases the flexibility to represent floorplans with complex inner structures. With the splitting strategy $(M \times N,h)$, we uniformly split each density region into $M \times N$ grids named the unit regions if the sliced grid has size larger than $h$. Otherwise, we progressively reduce the grid number to $M' \times N'$ until the grid size is larger than $h$. The unit region partition with different splitting strategies are shown in Figure~\ref{fig:unit_region}. We can see how 
an appropriate splitting strategy helps to maintain the alignment of the grids between adjacent density regions and achieves appropriate compactness by controlling the minimal grid size $h$.

The finer-level unit region set is highly consistent with the room partition based on the perceptual aesthetic rules. We examine the correlation between the unit regions and the room segmentations to confirm its effectiveness, reported in Table~\ref{tab:correlation_check}. Given a room segmentation map in the form of the $256 \times 256$ image, we assign the room label of each unit region as the voting result from all the pixels therein. The mean IoU between the region-wise and the pixel-wise label map reflects the consistency between them. The quantitative IoU evaluations indicate that an appropriate splitting strategy is important to represent the complex floorplan structures with a compact set of unit regions. 

\subsection{URE-Net architecture}

We train our network, named URE-Net, to learn the latent encoding of each unit region, as in the bottom row of Figure~\ref{fig:teaser}. Although defined on the sliced unit regions, the encoding representation should carry the information of the global geometric shape for the semantic requirement in many applications.

Our network is composed of two modules, a dense encoding module and an integration module, to obtain the encoding representation. In the dense encoding stage, we concatenate the rasterized floorplan mask and the normalized density map as input and utilize the Encoder-Decoder network to compute a dense pixel-wise feature map. We compute the inverse of the density value, filter with a Sobel operator, and divide it by 255 to obtain the normalized density map.
In addition to the floorplan mask, taking the normalized density map as an additional channel of the input will enable the network to learn from a more explicit geometric feature and thus achieve better semantic understanding. In the next stage, we integrate the local features within each unit region via Region pooling. We implement the region pooling of the local features within each unit region as RoI pooling and process using a MLP network with shared weights for each region. The output representation can be denoted as $R=\{f_1, f_2, ..., f_N\}$, for a floorplan divided into $N$ unit regions.

During training, the produced representation will be processed particularly for various applications and trained with the corresponding loss functions, as we will describe separately for each application in Section~\ref{sec:application}.

\section{Applications}
\label{sec:application}


\subsection{Interior Space Planning} 

The rooms in a modern residential house are often divided into several functional areas related with human activities, such as meeting, dining, entertainment zones, etc. There exist some common rules for the functional area partition, considering the space shape and size, relations to the windows and doors, relative position between these areas. For example, the living area is ususlly a relative large space with one or more windows. 
The interior space planning task is to divide the functional areas within a large room, which is close to a deterministic solution. Therefore, we formulate this task as a semantic segmentation problem.

\textbf{Network.}
In this application, the input is a multi-channel image of the room floorplan, where the channels are (i) the binary mask of floorplan interior space, (ii) N-valued masks of windows, doors, walls separately where the value indicating the room type they connect to, and (iii) the normalized density map as an additional channel. We use our URE-Net to compute the representation $R=\{f_1,f_2,...,f_N\}$ and then each region encoding is 
used to predict the functional area partition, i.e. the functionality label per region. The whole network is trained with the region-wise cross entropy loss on the predicted and ground-truth labels.


\begin{figure}[h]
  \centering
  \includegraphics[width=0.9\linewidth]{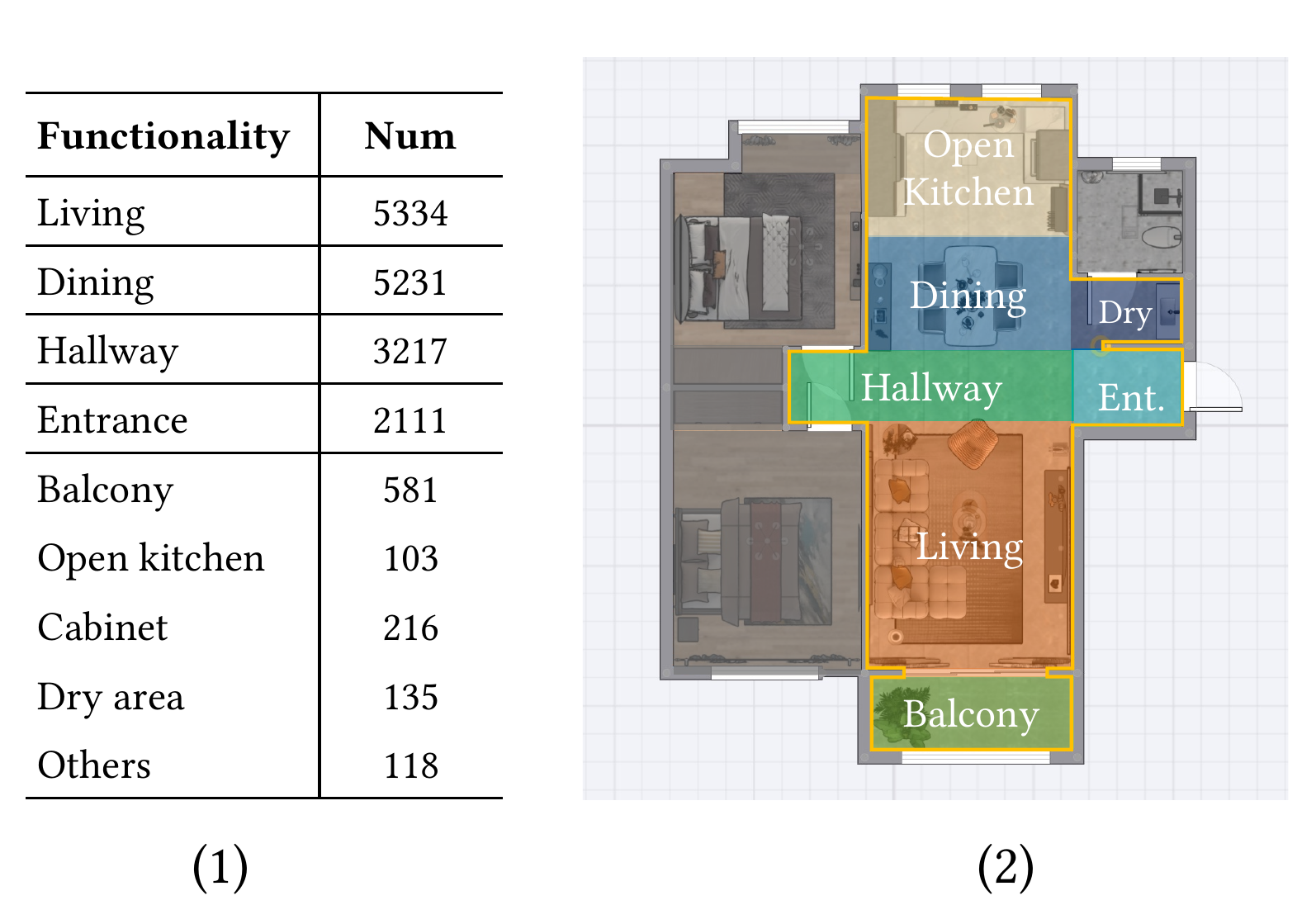}
  \caption{The statistics and a visualization of the collected functional area annotation on 3D-FRONT dataset~\cite{fu20213d}. We consider the four majority types and merge the rest of them as "others", forming five categories in our setting.}
  \Description{.}
  \label{fig:3d_front_data}
\end{figure}

\begin{figure}[h]
  \centering
  \includegraphics[width=\linewidth]{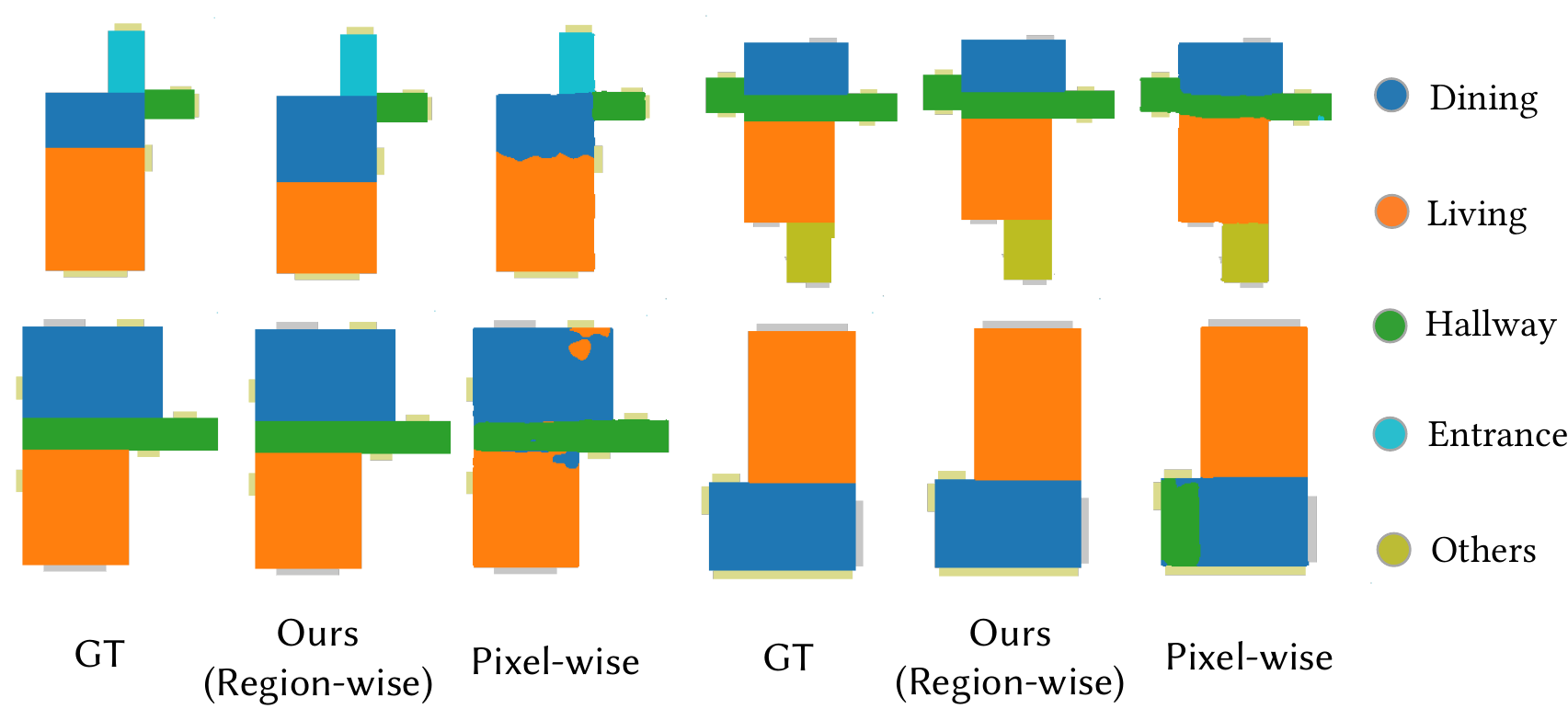}
  \caption{Qualitative comparison of the interior space planning application. The results are produced by DeepLabv3+ network (pixel-wise) and URE-Net+DeepLabv3+ (region-wise).}
  \Description{ttttttttttt.}
  \label{fig:space_plan_app_01}
\end{figure}

\textbf{Dataset.} Based on the 3D-FRONT dataset~\cite{fu20213d}, we collected a total of 5340 professional annotations of the functional area partition and perform random 80/20 split as the training and test set. The 9 types of functional areas are merged into 5 categories: (i) living, (ii) dining, (iii) entrance, (iv) hallway, and (v) others, as listed in Figure~\ref{fig:3d_front_data}-(1) 

  

\textbf{Metrics.} We adopt the Intersection over Union (IoU) per category and the average value among all the categories (mIoU) to measure the accuracy, as did in the other segmentation works. On the other hand, the boundary regularity of each functional area is usually preferred in realistic scenarios. Therefore, we compute the Boundary F-scores \cite{perazzi2016benchmark, gao2020beyond}, which measures the precision and recall between the predicted and ground-truth boundary with a pre-defined distance threshold. We reports results with threshold within 1 pixel for all boundaries and only internal boundaries, denoted as "All" and "Internal" respectively in Table~\ref{tab:space_planning_main1} and Table~\ref{tab:space_planning_ablation2}.


\textbf{Comparisons.} 
We consider the interior space planning as a semantic segmentation problem and compare with the state-of-the-art segmentation networks: DeepLabv3+ \cite{chen2018encoder}, HR-Net\cite{sun2019deep}, and U-Net\cite{ronneberger2015u}, under the same backbone settings. 
Specifically, we take the above networks for the pixel-wise segmentation as did in their papers and also utilize them as the dense encoding network of our URE-Net to make the comparison. The pixel-wise segmentations are retrained on our dataset with their default hyperparameter settings. As for the experiments to produce our representation, we use the unit region partition generated under the $(8 \times 8,1m)$ strategy. During training, we use Adam optimizer with a momentum of 0.9, a weight decay of 0.0005 and batch size of 100. The learning rate is set 5e-4 and decays 0.1 every 30 epochs and the optimization stops at 100 epochs.

 \begin{table*}[tp]
    \fontsize{6.5}{8}\selectfont  
    \centering  
   \caption{The IoU and Boundary F score of the interior space planning experiment results on 3D-FRONT dataset~\cite{fu20213d}. 1) The first group lists results using state-of-the-art methods on pixel-wise representation. 2) The rows in the second group list our URE-Net results using different network implementation under $(8\times 8\times1m)$ unit region partition. 3) The third group lists results under different unit region splitting strategies with DeepLabv3+ network.}
   \label{tab:space_planning_main1}
   \begin{tabular}{l|cccccc|cc}
 
     \toprule
     \multirow{3}*{Network} &
     \multicolumn{6}{c|}{IoU$(\uparrow)$} & \multicolumn{2}{c}{Boundary F score$(\uparrow)$} \\
     \cline{2-9}
      {} & Mean &Living&Dining& Hallway & Entrance & Others& All & Internal \\
     \midrule
     DeepLabv3+ & 70.06& 88.04& 77.66& 71.02& 58.24& 55.32& 60.58& 23.69 \\
     HR-Net & 69.29& 88.12& 77.82& 69.36& 57.96& 53.2& 60.69& 22.42 \\
     UNet & 68.04& 85.62& 73.52& 68.66& 60.8& 51.6& 59.06& 23.58 \\
     \midrule
     URE-Net+DeepLabv3+ &72.32& 89.94& 80.02& 73.64& 60.74& 57.24& 72.71& 63.19 \\
     URE-Net+HR-Net & 70.77& 89.5& 79.18& 72.92& 56.98& 55.28& 70.98& 61.85 \\
     URE-Net+UNet & 70.5& 87.42& 76.46& 72.96& 62.08& 53.58& 71.26& 61.67 \\
     \midrule
     $(1\times1,1.0m)$& 70.8& \textbf{90.26}& 74.42& \textbf{75.48}& 58.26& 55.58& 73.35& \textbf{68.17} \\
     $(2\times2,1.0m)$ & 71.9& 88.2& 78.6& 73.68& 62.1& 56.92& \textbf{73.62}& 66.34 \\
     $(4\times4,1.0m)$ & 72.05& 89.5& 79.12& 74.06& 61.74& 55.82& 72.85& 64.06 \\
     $(8\times8,1.0m)$ & \textbf{72.32}& 89.94& \textbf{80.02}& 73.64& 60.74& \textbf{57.24}& 72.71& 63.19 \\
     $(4\times4,0.5m)$ & 72.03& 89.12& 78.4& 74.12& \textbf{62.58}& 55.94& 71.03& 61.27 \\
     $(8\times8,0.5m)$ & 71.9& 89.72& 79.72& 73.24& 62.04& 54.76& 70.11& 58.98 \\

     \bottomrule
   \end{tabular}
 \end{table*}
 
  \begin{table*}[tp]
    \fontsize{6.5}{8}\selectfont  
    \centering  
  \caption{The IoU and Boundary F score of the interior space planning experiment results on 3D-FRONT dataset~\cite{fu20213d}. 1) The first group lists results using state-of-the-art methods on pixel-wise representation. 2) The rows in the second group list our URE-Net results using different network implementation under $(8\times 8\times1m)$ unit region partition. 3) The third group lists results under different unit region splitting strategies with DeepLabv3+ network.}
  \label{tab:space_planning_main122}
  \begin{tabular}{lc|cccccc|cc}
 
     \toprule
     \multirow{3}*{Network} &\multirow{3}*{Density Map} &
     \multicolumn{6}{c|}{IoU$(\uparrow)$} & \multicolumn{2}{c}{Boundary F score$(\uparrow)$} \\
     \cline{3-10}
      {} & {}&Mean &Living&Dining& Hallway & Entrance & Others& All & Internal \\
     \midrule
     DeepLabv3+ &\checkmark& 70.06& 88.04& 77.66& 71.02& 58.24& 55.32& 60.58& 23.69 \\
     HR-Net &\checkmark& 69.29& 88.12& 77.82& 69.36& 57.96& 53.2& 60.69& 22.42 \\
     UNet &\checkmark& 68.04& 85.62& 73.52& 68.66& 60.8& 51.6& 59.06& 23.58 \\
     \midrule
     URE-Net+DeepLabv3+ &\checkmark& \textbf{72.73}& \textbf{89.94}& \textbf{80.22}& \textbf{74.24}& 61.88& \textbf{57.36}& \textbf{72.82}& \textbf{63.49} \\
     pixelwise+voting &\checkmark& 72.62& 90.12& 80.38& 74.32& 62.14& 56.16& 73.56& 61.36 \\
     URE-Net+HR-Net &\checkmark& 70.77& 89.5& 79.18& 72.92& 56.98& 55.28& 70.98& 61.85 \\
     URE-Net+UNet &\checkmark& 70.5& 87.42& 76.46& 72.96& 62.08& 53.58& 71.26& 61.67 \\
     \midrule
     \multicolumn{10}{c}{Ablation: Region Partition}  \\
     \midrule
     $(1\times1,1.0m)$&\checkmark& 70.8& \textbf{90.26}& 74.42& \textbf{75.48}& 58.26& 55.58& 73.35& \textbf{68.17} \\
     $(2\times2,1.0m)$ &\checkmark& 71.9& 88.2& 78.6& 73.68& 62.1& 56.92& \textbf{73.62}& 66.34 \\
     $(4\times4,1.0m)$ &\checkmark& 72.05& 89.5& 79.12& 74.06& 61.74& 55.82& 72.85& 64.06 \\
     $(8\times8,1.0m)$ &\checkmark& \textbf{72.73}& 89.94& \textbf{80.22}& 74.24& 61.88& \textbf{57.36}& 72.82& 63.49 \\
     $(4\times4,0.5m)$ &\checkmark& 72.03& 89.12& 78.4& 74.12& \textbf{62.58}& 55.94& 71.03& 61.27 \\
     $(8\times8,0.5m)$ &\checkmark& 71.9& 89.72& 79.72& 73.24& 62.04& 54.76& 70.11& 58.98 \\
     \midrule
     \multicolumn{10}{c}{Ablation: Density Map}  \\
     \midrule
     DeepLabv3+ &-&  69.54& 88.22& 78.04& 69.78& 57.76& 53.92& 59.77& 19.63 \\
     DeepLabv3+ &\checkmark& \textbf{70.06}& 88.04& 77.66& \textbf{71.02}& \textbf{58.24}& \textbf{55.32}& \textbf{60.58}& \textbf{23.69} \\
     \midrule
     URE-Net+DeepLabv3+ &-& 71.6& 89.8& 79.78& 73.7& 59.12& 55.58& 71.97& 62.72\\
     URE-Net+DeepLabv3+ &\checkmark& \textbf{72.73}& \textbf{89.94}& \textbf{80.22}& \textbf{74.24}& \textbf{61.88}& \textbf{57.36}& \textbf{72.82}& \textbf{63.49} \\
     \bottomrule
  \end{tabular}
 \end{table*}

\begin{table}[tp]
  \fontsize{6.5}{8}\selectfont  
  \centering  
 \caption{Ablation results of the interior space planning application to validate the effect of density map. The original DeepLabv3+ network is designed to output the pixel-wise segmentation while our URE-Net produces region-wise segmentation.}
 \label{tab:space_planning_ablation2}
 \begin{tabular}{lc|c|cc}

   \toprule
   \multirow{3}*{Network} & \multirow{3}*{Density} &
   \multirow{3}*{mIoU$(\uparrow)$} & \multicolumn{2}{c}{Boundary F score$(\uparrow)$} \\
   \cline{4-5}
    {} & {}& {} & All& Internal \\
   \midrule
   DeepLabv3+ & - &69.54 &59.77& 19.63 \\
   DeepLabv3+ & \checkmark &\textbf{70.06} &\textbf{60.58}& \textbf{23.69}  \\
   \midrule
   URE-Net &  - &71.60 &71.97& 62.72  \\
   URE-Net & \checkmark &\textbf{72.32} &\textbf{72.71}& \textbf{63.19} \\

   \bottomrule
 \end{tabular}
\end{table}


Table~\ref{tab:space_planning_main1} shows the IoU and Boundary F scores for the comparison. Our Unit Region Encoding constantly outperforms the pixel-wise segmentation with the corresponding network architecture. The IoU accuracy benefits from the region-wise encodings. More importantly, the higher Boundary F scores indicate that our approach is able to construct a more highly regular functional area segmentation (see Figure~\ref{fig:space_plan_app_01}) without the tedious post-processing steps.


\textbf{Ablation Study.} We use the URE-Net with DeepLabv3+ network architecture, which achieves the best performance in the comparison experiments, to conduct the ablation study. The goal is to understand the effect of splitting strategy and the geometry-aware density map.

We first train our network under different splitting strategies: $(1\times1, 1.0m)$, $(2\times2, 1.0m)$, $(4\times4, 1.0m)$, $(4\times4, 0.5m)$, $(8\times8, 1.0m)$, $(8\times8, 0.5m)$. From Table~\ref{tab:space_planning_main1}, we observe that the strategy $(8 \times 8, 1m)$ obtains the best mean IoU accuracy. A too coarse partition lacks of enough flexibility to represent multiple functional areas, while a too fine-grained partition will lose the advantage of the unit region partition which is aware of the complex floorplan boundaries. However, it's worth noting that the different splitting strategies always obtain better performance than the pixel-wise segmentation.

We also compare the experiment results with and without normalized density map as the input channel, as reported in Table~\ref{tab:space_planning_ablation2}. The normalized density map brings 0.52 and 0.72 mIoU increase for the pixel-wise segmentation and region-wise segmentation respectively. It also improves the Boundary F score due to its regularity.


\subsection{Floorplan Metric Learning} 


The metric learning problem is to learn the similarity metric between floorplans with complete room partition and annotated labels, which is fundamental for the floorplan classification and retrieval applications.
We follow the common practice to map the floorplans into embedding vectors from our representation and compute their Euclidean distance as the similarity measurement.

\textbf{Network.} Since the room partition is already given in the input floorplan, we take the rooms as our unit regions and use URE-Net to learn the region-wise latent encodings.
Our floorplan representation $R=\{f_1,f_2,...,f_N\}$ is the features defined on a set of unordered unit regions with varing number of regions. Inspired by the solution in PointNet \cite{qi2017pointnet}, we adopt their feature transform and element-wise max pooling function as the symmetric function $g$ to obtain the embedding vector of the entire floorplan:
\begin{equation}
f_g=g([f_1,f_2,...,f_N]).
\end{equation}
The entire network is trained on the triplet loss defined on the sampled triplets in the training set. Following the protocol of LayoutGMN \cite{patil2021layoutgmn}, we take their training set built on the large-scale RPLAN dataset \cite{wu2019data} which is composed of the sampled triplets of floorplans. Each triplet contains an anchor floorplan with the positive and negative samples from the dataset. The performance is evaluated on the test triplets by comparing to the IoU-based ground-truth as well as the provided user annotations.



\textbf{Compairons.} We compare our method with four baselines, Graph Kernel(GK)~\cite{fisher2011characterizing}, U-Net~\cite{ronneberger2015u}, GCN-CNN~\cite{manandhar2020learning}, and LayoutGMN~\cite{patil2021layoutgmn}. Since the training setting is exactly the same with LayoutGMN, we take the numbers from their paper and present our results. Our URE-Net is trained with Adam optimizer with a momentum of 0.9, a weight decay of 0.0005 and batch size of 10. The learning rate is 1e-4 and decays 0.8 every 20 epochs for a total of 200 epochs.

\begin{figure}[h]
  \centering
  \includegraphics[width=\linewidth]{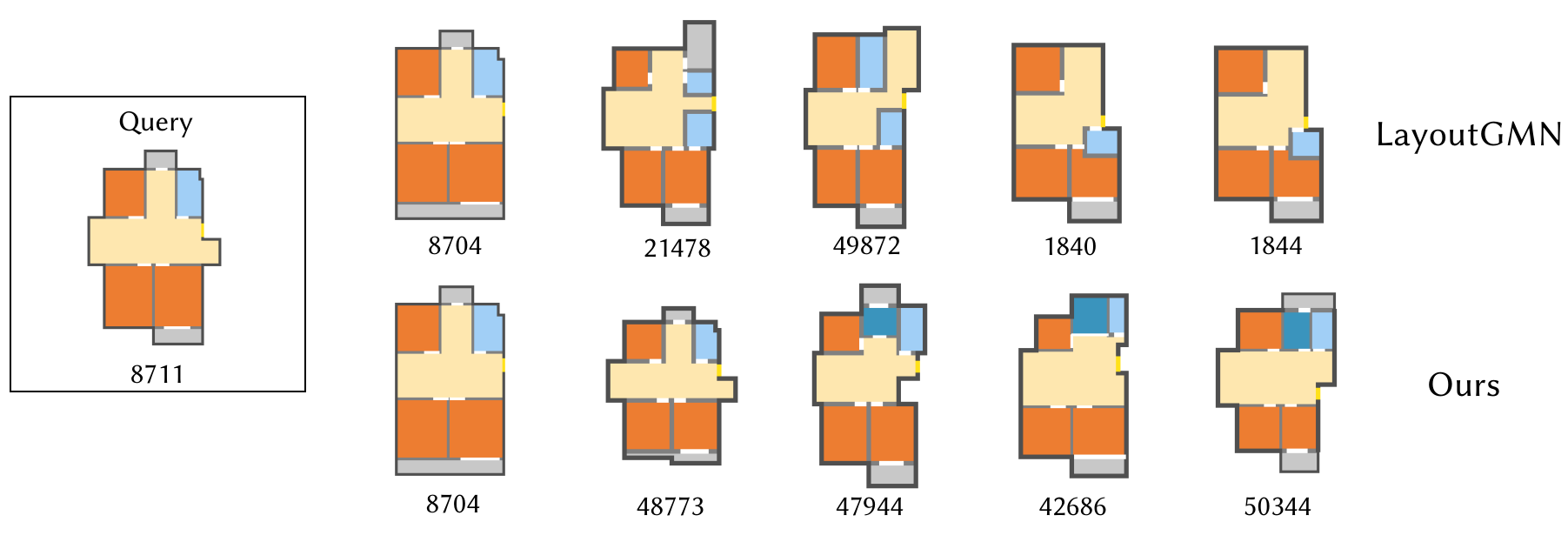}
  \caption{The floorplan retrieval comparison between our method and LayoutGMN~\cite{patil2021layoutgmn}.}
  \Description{ttttttttttt.}
  \label{fig:app_02}
\end{figure}

\begin{table}[tp]
    \fontsize{6.5}{8}\selectfont  
    \centering  
   \caption{Quantitative evaluation of the metric learning comparisons.}
   \label{tab:retrieval}
   \begin{tabular}{l|c|c}
  
     \toprule
     \multirow{3}*{Method} & 
      \multicolumn{2}{c}{Method Test Accuracy on Triplets} \\
     \cline{2-3}
      {} &IoU-based$(\uparrow)$& User-based$(\uparrow)$ \\
     \midrule
     Graph Kernel & 92.07 &95.60  \\
     U-Net Triplet  & 93.01 &91.00   \\
     GCN-CNN Triplet  & 92.50 &91.80   \\
     LayoutGMN & 97.54 &97.60  \\
     \midrule
      URE-Net w/o density & 99.3 &98.5\\
      URE-Net w/ density & \textbf{99.48} &\textbf{98.62}\\
     \bottomrule
   \end{tabular}
  \end{table}

Table~\ref{tab:retrieval} validates that our method outperforms all the other state-of-the-art methods on the accuracy performance. The aforementioned methods are based on either graph structures with hand-crafted geometric features for the room nodes and edges (i.e. GK, LayoutGMN), or the rasterized semantic images lacking of structural room partitions (U-Net). In contrast, our method in fact plays a \emph{graph on image} function where we integrate the dense feature map learned from the image into room-wise encodings. The learned encodings carry more shape information than the hand-crafted features, while the room-wise pooling enables the awareness of room partitions. We can also observe the accuracy increase brought by the input normalized density map with its dense geometric information. In addition, we also present the qualitative comparison with LayoutGMN on the floorplan retrieval task in Figure~\ref{fig:app_02}.

\subsection{Floorplan Generation.} 

\begin{figure}[h]
  \centering
  \includegraphics[width=\linewidth]{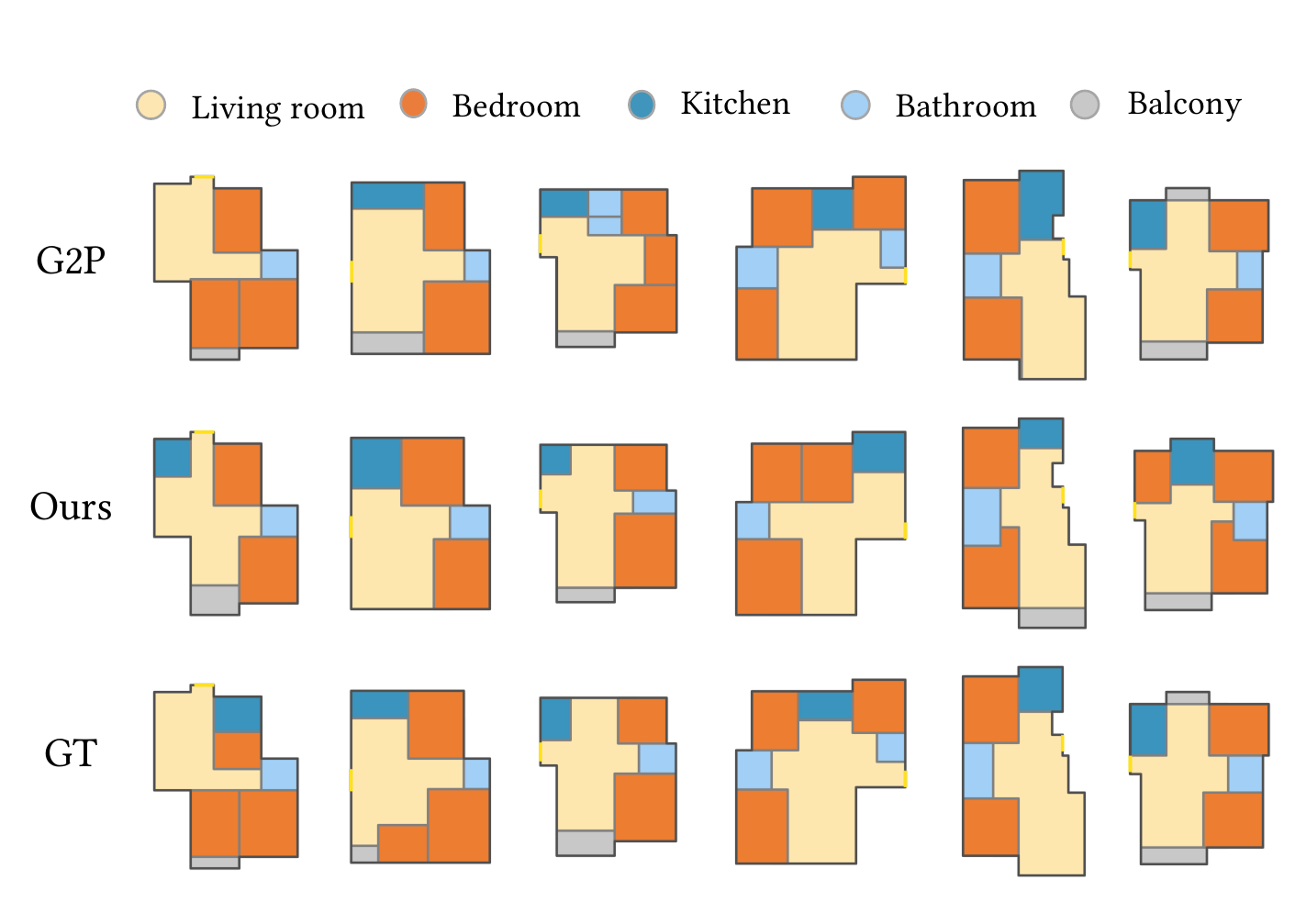}
  \caption{The ground-truth floorplan and the generated results of our method and Graph2Plan~\cite{hu2020graph2plan}.}
  \Description{ttttttttttt.}
  \label{fig:generate_app_01}
\end{figure}

\begin{figure}[h]
  \centering
  \includegraphics[width=\linewidth]{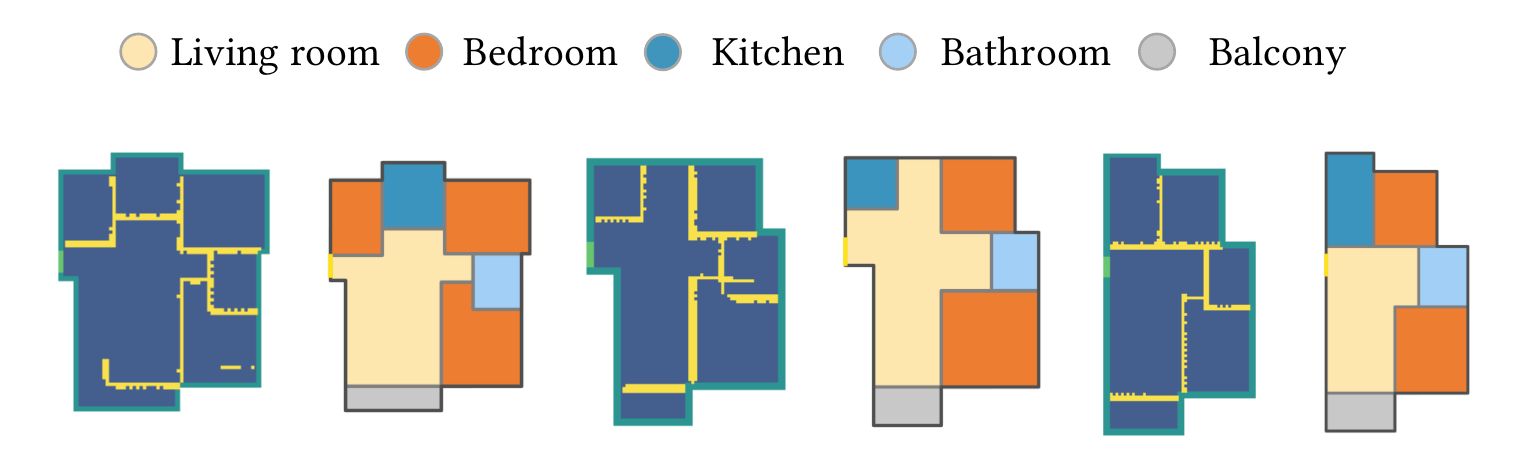}
  \caption{The predicted wall masks and the generation results of our method.}
  \Description{ttttttttttt.}
  \label{fig:generate_app_02}
\end{figure}

The floorplan generation application is to automatically generate the room partitions and the room labels from an input empty floorplan boundary. Comparing to the interior space planning, the floorplan generation task is more challenging and non-deterministic with the complex room partition of the entire floorplan rather than coarse functional areas within a large room.

\textbf{Network.}
We propose a two-step pipeline to produce plausible and high-quality floorplans. In the first step, we train a generative Pix2Pix network\cite{isola2017image} to output a binary wall mask image. The input is the N-valued floorplan mask, where the labels $0,1,2,3$ denote the exterior space, interior space, exterior wall, front door respectively, as well as the normalized density map concatenated to form a two-channel image. Although the output wall masks are not completed, full of burrs, and redundant, it provides the proper guidance for the complete floorplan generation. In the next step, we make use of our Unit Region Encoding to segment and label the rooms. Specifically, taking the input of the first stage and the generated wall mask image as input, our URE-Net extracts the latent encoding of each unit region. We compute the Euclidean distance between the region features to measure their closeness: the distance smaller than a pre-defined threshold indicates that the two unit regions should belong to the same room. Additionally, we predict the room label from each region encoding and vote to determine the type of the room. Please note that our second step actually performs a one-stage instance segmentation\cite{wang2018sgpn} to construct the room partition with regular shapes without post-processing vectorization.


The training process contains three steps. In the first step, we train the wall generation network on the annotations from the training set for 100 epochs with their default setting. Then in the next step, we train the URE-Net with DeepLabv3+ architecture 100 epochs to learn room instance segmentation from the ground-truth room partition and labels, under the $(8 \times 8, 1m)$ splitting strategy. The loss function is a combination of LS-GAN loss~\cite{mao2017least} and L2 reconstruction loss. Note that this step takes the ground-truth wall masks as input for a more stable training. In the above two steps, the learning rate is 0.0002 and decays 0.8 every 20 epochs. Finally, we fine-tune the URE-Net to help the network with the generated wall masks as input to reduce the domain gap between the generated and the ground-truth wall masks. During fine-tuning, we randomly replace half of the ground-truth wall masks by the generated wall masks. The initial learning rate of fine-tuning is 0.0001 and linearly decay to zero over 50 epochs. For each above steps, we use Adam optimizer with momentum of 0.9 and batch size 64.

\textbf{Dataset.} We conduct the experiment on the commonly used RPLAN dataset\cite{wu2019data}, which contains 13 types of rooms. We merge the room categories into 5 types: \emph{LivingRoom}, \emph{Bedroom}, \emph{Kitchen}, \emph{Bathroom} and \emph{Balcony}, since it's hard to distinguish some of them, like the MasterRoom and SecondRoom. The dataset is randomly split into 75K training set and 5K test set.



\textbf{Results.} We show the qualitative comparison in Figure~\ref{fig:generate_app_01} and more of our results in~\ref{fig:generate_app_02}. To the best of our knowledge, there's no widely adopted metric to measure the plausibility of the generated floorplans. Therefore, we perform the user study, which is the common practice in the related works \cite{wu2019data, hu2020graph2plan}. We recruit 30 professional designers and organize the floorplans from different sources (ground-truth, our results, the results of Graph2Plan~\cite{hu2020graph2plan}) into pairs. We present the floorplan pairs to the designers in a random order and ask them to select the more plausible one or "cannot tell". Each volunteer is involved in two groups of comparisons: The first group is 20 pairs of randomly selected our generated floorplans and the corresponding ground-truth floorplans with the same exterior boundaries. The second group contains 20 pairs of generated floorplans from our result and Graph2Plan from same exterior boundary inputs from the test set.


The 600 collected votes for the options "GT/ours/cannot tell" are 326/237/37 respectively, and the other 600 votes for the options "Graph2Plan/ours/cannot tell" are 273/232/95. It indicates that our generated floorplans are close to the ground-truth and are comparable to the state-of-the-art Graph2Plan method, while our method is more concise with a simple two-step pipeline without tedious post-processing steps. 




\vspace{-0.4cm}
\section{Conclusion}
In this work, we propose to learn the Unit Region Encoding representation defined as a set of latent encoding features on a unit region partition, which can be flexibility utilized in many different applications. We define the geometric-aware density map, which is used to divide the boundary-adaptive compact unit region partition and processed as input for the learning of the encoding features. The extensive comparison and ablation study experiments validate the superiority of the representation with our unit region partition and the density map.

Although the use of patch-wise or region-wise features has been adopted in many computer vision works such as ViT~\cite{Dosovitskiy2021AnII} we argue that the region partition should be carefully designed for the particular data domains, especially for floorplans which have highly irregular and complex boundary shapes. Following the idea of our Unit Region Encoding, there are many interesting directions for further exploration. For example, similar to the use of the normalized density map, learning the encoding features from the input floorplan masks containing more channels reflecting the human trajectory or environment lighting information will improve the floorplan generation and planning tasks. On the other hand, it is also worth seeking a more powerful network architecture to produce structural encoding representation rather than the current discrete set of latent encodings.

\bibliographystyle{ACM-Reference-Format}
\bibliography{URF-acmtog}
\end{document}